\documentclass[journal]{IEEEtran}

\usepackage{cite}

\ifCLASSINFOpdf
\else
  \usepackage{graphicx}
\fi

\usepackage{threeparttable}

\usepackage{psgo}
\setgounit{0.4cm}

\usepackage{soul}

\newcommand{\rv}[1]{#1}

\newcommand{\rvv}[1]{#1}

\newcommand{\rvvv}[1]{#1}

\newcommand{\rvvvv}[1]{#1}

\usepackage{algorithm}
\usepackage{algorithmic}

\usepackage{amsmath}
\usepackage{url}

\begin{document}
%
\title{On Move Pattern Trends\\in \rvv{a} Large Go Games Corpus}

\author{Petr~Baudi\v{s},~Josef~Moud\v{r}\'{i}k\\
September 2012
\thanks{P. Baudi\v{s} is student at the Faculty of Math and Physics, Charles University, Prague, CZ, and also does some of his Computer Go research as an employee of SUSE Labs Prague, Novell CZ.}
\thanks{J. Moud\v{r}\'{i}k is student at the Faculty of Math and Physics, Charles University, Prague, CZ.}}

%
%

\markboth {On Move Pattern Trends in Large Go Games Corpus}{}

%



\maketitle

\begin{abstract}

We process a~large corpus of game records of the board game of Go and propose
a~way of extracting summary information on played moves. We then apply several
basic data-mining methods on the summary information to identify the most
differentiating features within the summary information, and discuss their
correspondence with traditional Go knowledge. We show statistically significant
mappings of the features to player attributes such as playing strength or
informally perceived ``playing style'' (e.g. territoriality or aggressivity),
describe accurate classifiers for these attributes, and propose applications
including seeding real-work ranks of internet players, aiding in Go study and
tuning of Go-playing programs, or contribution to Go-theoretical discussion on
the scope of ``playing style''.

\end{abstract}


%
\IEEEpeerreviewmaketitle


\section{Introduction}
%
%
%
%
\IEEEPARstart{T}{he} field of Computer Go usually focuses on the problem
of creating a~program to play the game, finding the best move from a~given
board position \cite{GellySilver2008}.
We will make use of one method developed in the course
of such research and apply it to the analysis of existing game records
with the aim of helping humans to play and understand the game better
instead.

Go is a~two-player full-information board game played
on a~square grid (usually $19\times19$ lines) with black and white
stones; the goal of the game is to surround the most territory and
capture enemy stones. We assume basic familiarity with the game.

Many Go players are eager to play using computers (usually over
the internet) and review games played by others on computers as well.
This means that large amounts of game records are collected and digitally
stored, enabling easy processing of such collections. However, so far
only little has been done with the available data. We are aware
only of uses for simple win/loss statistics \cite{KGSAnalytics} \cite{ProGoR}
and ``next move'' statistics on a~specific position \cite{Kombilo} \cite{MoyoGo}.

\rvvv{Additionally}, \rvv{a simple machine learning technique based on GNU Go's}\cite{GnuGo}\rvv{
move evaluation feature has recently been presented in}\cite{CompAwar}\rvv{. The authors used decision trees
to predict whether a given user belongs into one of three classes based on his strength
(causal, intermediate or advanced player). This method is however limited by the
blackbox-use of GNU Go engine, making it unsuitable for more detailed analysis of the moves.}

We present a~more in-depth approach --- from all played moves, we devise
a~compact evaluation of each player. We then explore correlations between
evaluations of various players in the light of externally given information.
This way, we can discover similarity between move characteristics of
players with the same playing strength, or discuss the meaning of the
``playing style'' concept on the assumption that similar playing styles
should yield similar move characteristics.

\rv{We show that a~sample of player's games can be used to quite reliably estimate player's strength,
game style, or even a time when he/she was active. Apart from these practical results,
the research may prove to be useful for Go theoretists by investigating the principles behind
the classical ``style'' classification.}


\rv{We shall first present details of the extraction and summarization of
information from the game corpus (section~}\ref{pattern-vectors}\rv{).
Afterwards, we will explain the statistical methods applied (section~}\ref{data-mining}\rv{),
and then describe our findings on particular game collections,
regarding the analysis of either strength (section~}\ref{strength-analysis}\rv{)
or playing styles (section~}\ref{style-analysis}\rv{).
Finally, we will explore possible interpretations and few applications
of our research (section~}\ref{proposed-apps-and-discussion}\rv{)
and point out some possible future research directions (section~}\ref{future-research}\rv{).}

\section{Data Extraction}
\label{pattern-vectors}

As the input of our analysis, we use large collections of game records
\rvv{in SGF format} \cite{SGF}
\rv{grouped by the primary object of analysis
(player name when analyzing style of a particular player,
player rank when looking at the effect of rank on data, etc.).}
We process the games, generating a description for each
played move -- a {\em pattern}, being a combination of several
{\em pattern features} described below.

We \rv{compute the occurence counts of all encountered patterns,
eventually} composing $n$-dimensional {\em pattern vector}
$\vec p$ of counts of the $n$ \rvv{(we use $n = 500$)} globally most frequent patterns
(the mapping from patterns to vector elements is common for
\rv{all generated vectors}).
We can then process and compare just the pattern vectors.

\subsection{Pattern Features}
When deciding how to compose the patterns we use to describe moves,
we need to consider a specificity tradeoff --- overly general descriptions carry too few
information to discern various player attributes; too specific descriptions
gather too few specimen over the games sample and the vector differences are
not statistically significant.

We have chosen an intuitive and simple approach inspired by pattern features
used when computing Elo ratings for candidate patterns in Computer Go play
\cite{PatElo}. Each pattern is a~combination of several {\em pattern features}
(name--value pairs) matched at the position of the played move.
We use these features:

\begin{itemize}
\item capture move flag\rvvv{,}
\item atari move flag\rvvv{,}
\item atari escape flag\rvvv{,}
\item contiguity-to-last flag%
\footnote{We do not consider contiguity features in some cases when we are working
on small game samples and need to reduce pattern diversity.}
--- whether the move has been played in one of 8 neighbors of the last move\rvvv{,}
\item contiguity-to-second-last flag\rvvv{,}
\item board edge distance --- only up to distance 4\rvvv{,}
\item and spatial pattern --- configuration of stones around the played move.
\end{itemize}

The spatial patterns are normalized (using a dictionary) to be always
black-to-play and maintain translational and rotational symmetry.
Configurations of radius between 2 and 9 in the gridcular metric%
\footnote{The {\em gridcular} metric
$d(x,y) = |\delta x| + |\delta y| + \max(|\delta x|, |\delta y|)$ produces
a circle-like structure on the Go board square grid \cite{SpatPat}. }
are matched.

Pattern vectors representing these features contain information on
played shape as well as a basic representation of tactical dynamics
--- threats to capture stones, replying to last move, or ignoring
opponent's move elsewhere to return to an urgent local situation.
The shapes \rv{often} correspond to opening moves
(either in empty corners and sides, or as part of {\em joseki}
--- commonly played sequences) characteristic for a certain
strategic aim. In the opening, even a single-line difference
in the distance from the border can have dramatic impact on
further local and global development.

\subsection{Vector Rescaling}

The pattern vector elements can have diverse values since for each object,
we consider a different number of games (and thus patterns).
Therefore, we normalize the values to range $[-1,1]$,
the most frequent pattern having the value of $1$ and the least occuring
one being $-1$.
Thus, we obtain vectors describing relative frequency of played patterns
independent on number of gathered patterns.
But there are multiple ways to approach the normalization.

\begin{figure}[!t]
\centering
\includegraphics{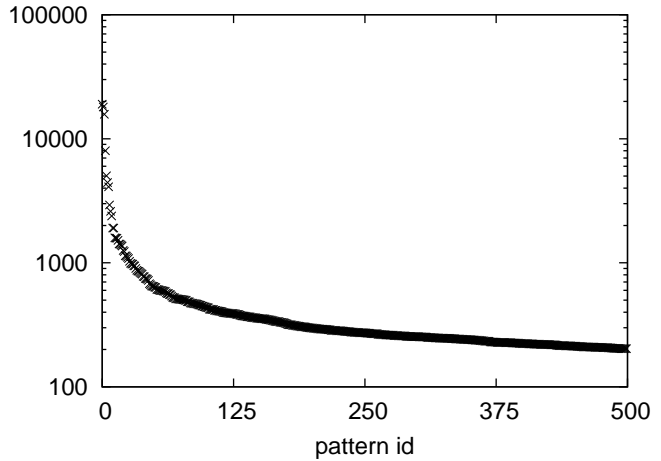}
\caption{Log-scaled number of pattern occurences
in the GoGoD games examined in sec. \ref{style-analysis}.}
\label{fig:patcountdist}
\end{figure}

\subsubsection{Linear Normalization}

\rvv{An intuitive solution is to} linearly re-scale the values using:
$$y_i = {x_i - x_{\rm min} \over x_{\rm max}}$$
This is the default approach; we have used data processed by only this
computation unless we note otherwise.
As shown on fig. \ref{fig:patcountdist}, most of the spectrum is covered
by the few most-occuring patterns (describing mostly large-diameter
shapes from the game opening). This means that most patterns will be
always represented by only very small values near the lower bound.

\subsubsection{Extended Normalization}
\label{xnorm}

To alleviate this problem, we have also tried to modify the linear
normalization by applying two steps --- {\em pre-processing}
the raw counts using
$$x_i' = \log (x_i + 1)$$
and {\em post-processing} the re-scaled values by the logistic function:
$$y_i' = {2 \over 1 + e^{-cy_i}}-1$$
However, we have found that this method is not universally beneficial.
In our styles case study (sec. \ref{style-analysis}), this normalization
produced PCA decomposition with significant dimensions corresponding
better to some of the prior knowledge and more instructive for manual
inspection, but ultimately worsened accuracy of our classifiers.
\rvvv{From this we conjecture} that the most frequently occuring patterns are
also most important for classification of major style aspects.

\subsection{Implementation}

We have implemented the data extraction by making use of the pattern
features matching implementation
within the Pachi Go-playing program \cite{Pachi}, \rvv{which works according to 
the Elo-rating pattern selection scheme} \cite{PatElo}.
We extract information on players by converting the SGF game
records to GTP stream \cite{GTP} that feeds Pachi's {\tt patternscan}
engine, \rv{producing} a~single {\em patternspec} (string representation
of the particular pattern features combination) per move. Of course,
only moves played by the appropriate \rv{player} are collected.

\section{Data Mining}
\label{data-mining}

To assess the properties of gathered pattern vectors
and their influence on playing styles,
we analyze the data using several basic data minining techniques.
The first two methods {\em (analytic)} rely purely on single data set
and serve to show internal structure and correlations within the data set.

\rvvv{\emph{ Principal Component Analysis}} \rvv{\emph{(PCA)}} \cite{Jolliffe1986}
finds orthogonal vector components that \rv{represent} the largest variance
\rv{of values within the dataset.
That is, PCA will produce vectors representing
the overall variability within the dataset --- the first vector representing
the ``primary axis'' of the dataset, the next vectors representing the less
significant axes; each vector has an associated number that
determines its impact on the overall dataset variance: $1.0$ would mean
that all points within the dataset lie on this vector, value close to zero
would mean that removing this dimension would have little effect on the
overall shape of the dataset.}

\rvv{Reversing the process of the PCA by backprojecting the orthogonal vector components into the
original pattern space can indicate which patterns correlate with each component.}
Additionally, PCA can be used as vector preprocessing for methods
that are negatively sensitive to pattern vector component correlations.

\rv{On the other hand,} Sociomaps \cite{Sociomaps} \cite{TeamProf} \cite{SociomapsPersonal} produce
spatial representation of the data set elements (e.g. players) based on
similarity of their data set features\rvvv{. Projecting some other
information on this map helps illustrate connections within the data set.}


Furthermore, we test several \emph{classification} methods that assign
an \emph{output vector} $\vec O$ \rv{to} each pattern vector $\vec P$,
\rv{the output vector representing the information we want to infer
from the game sample} --- e.g.~\rv{assessment of} the playing style,
player's strength or even meta-information like the player's era
or the country of origin.
Initially, the methods must be calibrated (trained) on some prior knowledge,
usually in the form of \emph{reference pairs} of pattern vectors
and the associated output vectors.
The reference set is divided into training and testing pairs
and the methods can be compared by the mean square error (MSE) on testing data set
(difference of output vectors approximated by the method and their real desired value).


The most trivial method is approximation by the PCA representation
matrix, provided that the PCA dimensions have some already well-defined
\rv{interpretation}\rvvv{. This} can be true for single-dimensional information like
the playing strength.
Aside of that, we test the $k$-Nearest Neighbors (\emph{$k$-NN}) classifier \cite{CoverHart1967} 
that approximates $\vec O$ by composing the output vectors
of $k$ reference pattern vectors closest to $\vec P$.

Another classifier is a~multi-layer feed-forward Artificial Neural Network \rv{(see e.g. }\cite{Haykin1994}\rv{)}.
The neural network can learn correlations between input and output vectors
and generalize the ``knowledge'' to unknown vectors\rvvv{. The neural network} can be more flexible
in the interpretation of different pattern vector elements and discern more
complex relations than the $k$-NN classifier,
but may not be as stable and expects larger training sample.

Finally, a commonly used classifier in statistical inference is
the Naive Bayes Classifier \cite{Bayes}\rvvv{. It} can infer relative probability of membership
in various classes based on previous evidence (training patterns).

\subsection{Statistical Methods}
We use couple of general statistical analysis \rv{methods} together
with the particular techniques.
\label{pearson}
To find correlations within or between extracted data and
some prior knowledge (player rank, style vector), we compute the well-known
{\em Pearson product-moment correlation coefficient (PMCC)} \cite{Pearson},
measuring the strength of the linear dependence%
\footnote{A desirable property of PMCC is that it is invariant to translations and rescaling
of the vectors.}
between any two dimensions:

$$ r_{X,Y} = {{\rm cov}(X,Y) \over \sigma_X \sigma_Y} $$

To compare classifier performance on the reference data, we employ
{\em $k$-fold cross validation}:
we randomly divide the training set
into $k$ distinct segments of similar sizes and then iteratively
use each part as a~testing set as the other $k-1$ parts are used as a~training set.
We then average results over all iterations.

\subsection{Principal Component Analysis}
\label{PCA}
We use Principal Component Analysis 
to reduce the dimensions of the pattern vectors while preserving
as much information as possible, assuming inter-dependencies between
pattern vector dimensions are linear.
\rv{Technically}, PCA is an eigenvalue decomposition of a~covariance matrix of centered pattern vectors,
producing a~linear mapping $o$ from $n$-dimensional vector space
to a~reduced $m$-dimensional vector space.
The $m$ eigenvectors of the original vectors' covariance matrix
with the largest eigenvalues are used as the base of the reduced vector space;
the eigenvectors form projection matrix $W$.

For each original pattern vector $\vec p_i$,
we obtain its new representation $\vec r_i$ in the PCA base
as shown in the following equation:
\begin{equation}
\vec r_i = W \cdot \vec p_i
\end{equation}

The whole process is described in the Algorithm \ref{alg:pca}.

\begin{algorithm}
\caption{PCA -- Principal Component Analysis}
\begin{algorithmic}
\label{alg:pca}
\REQUIRE{$m > 0$, set of players $R$ with pattern vectors $p_r$}
\STATE $\vec \mu \leftarrow 1/|R| \cdot \sum_{r \in R}{\vec p_r}$
\FOR{ $r \in R$}
\STATE $\vec p_r \leftarrow \vec p_r - \vec \mu$
\ENDFOR
\FOR{ $(i,j) \in \{1,... ,n\} \times \{1,... ,n\}$}
\STATE $\mathit{Cov}[i,j] \leftarrow 1/|R| \cdot \sum_{r \in R}{\vec p_{ri} \cdot \vec p_{rj}}$
\ENDFOR
\STATE Compute Eigenvalue Decomposition of $\mathit{Cov}$ matrix
\STATE Get $m$ largest eigenvalues
\STATE Most significant eigenvectors ordered by decreasing eigenvalues form the rows of matrix $W$
\FOR{ $r \in R$}
\STATE $\vec r_r\leftarrow W \vec p_r$
\ENDFOR
\end{algorithmic}
\end{algorithm}

\subsection{Sociomaps}
\label{soc}
Sociomaps are a general mechanism for \rv{visualizing}
relationships on a 2D plane such that \rv{given} ordering of the
\rv{player} distances in the dataset is preserved in distances on the plane.
In our particular case,
we will consider a dataset $\vec S$ of small-dimensional
vectors $\vec s_i$. First, we estimate the {\em significance}
of difference {\rv of} each two subjects.
Then, we determine projection $\varphi$ of all the $\vec s_i$
to spatial coordinates of an Euclidean plane, such that it reflects
the estimated difference significances.


To quantify the differences between the subjects ({\em team profiling})
into an $A$ matrix, for each two subjects $i, j$ we compute the scalar distance%
\footnote{We use the {\em Manhattan} metric $d(x,y) = \sum_i |x_i - y_i|$.}
of $s_i, s_j$ and then estimate the $A_{ij}$ probability of at least such distance
occuring in uniformly-distributed input (the higher the probability, the more
significant and therefore important to preserve the difference is).

To visualize the quantified differences, we need to find
the $\varphi$ projection such that it maximizes a {\em three-way ordering} criterion:
ordering of any three members within $A$ and on the plane
(by Euclidean metric) must be the same.

$$ \max_\varphi \sum_{i\ne j\ne k} \Phi(\varphi, i, j, k) $$
$$ \Phi(\varphi, i, j, k) = \begin{cases}
	1 & \delta(1,A_{ij},A_{ik}) = \delta(\varphi(i),\varphi(j),\varphi(k)) \\
	0 & \hbox{otherwise} \end{cases} $$
$$ \delta(a, b, c) = \begin{cases}
	1  & |a-b| > |a-c| \\
	0  & |a-b| = |a-c| \\
	-1 & |a-b| < |a-c| \end{cases} $$

The $\varphi$ projection is then determined by randomly initializing
the position of each subject and then employing gradient descent methods.

\subsection{k-Nearest Neighbors Classifier}
\label{knn}
Our goal is to approximate \rvv{the} player's output vector $\vec O$,
knowing their pattern vector $\vec P$.
We further assume that similarities in players' pattern vectors
uniformly correlate with similarities in players' output vectors.

We require a set of reference players $R$ with known \emph{pattern vectors} $\vec p_r$
and \emph{output vectors} $\vec o_r$.
$\vec O$ is approximated as weighted average of \emph{output vectors}
$\vec o_i$ of $k$ players with \emph{pattern vectors} $\vec p_i$ closest to $\vec P$.
This is illustrated in the Algorithm \ref{alg:knn}.
Note that the weight is a function of distance and is not explicitly defined in Algorithm \ref{alg:knn}.
During our research, exponentially decreasing weight has proven to be sufficient\rvvv{,
as detailed in each of the case studies.}

\begin{algorithm}
\caption{k-Nearest Neighbors}
\begin{algorithmic}
\label{alg:knn}
\REQUIRE{pattern vector $\vec P$, $k > 0$, set of reference players $R$}
\FORALL{$r \in R$ }
\STATE $D[r] \leftarrow \mathit{EuclideanDistance}(\vec p_r, \vec P)$
\ENDFOR
\STATE $N \leftarrow \mathit{SelectSmallest}(k, R, D)$
\STATE $\vec O \leftarrow \vec 0$
\FORALL{$r \in N $}
\STATE $\vec O \leftarrow \vec O + \mathit{Weight}(D[r]) \cdot \vec o_r $
\ENDFOR
\end{algorithmic}
\end{algorithm}

\subsection{Neural Network Classifier}
\label{neural-net}

Feed-forward neural networks are known for their ability to generalize
and find correlations between input patterns and output classifications.
Before use, the network is iteratively trained on the training data
until the error on the training set is reasonably small.


\subsubsection{Computation and activation of the NN}
Technically, the neural network is a network of interconnected
computational units called neurons.
A feed-forward neural network has a layered topology\rvvv{. It}
usually has one \emph{input layer}, one \emph{output layer}
and an arbitrary number of \emph{hidden layers} between.
Each neuron $i$ gets input from all neurons in the previous layer,
each connection having specific weight $w_{ij}$.

The computation proceeds in discrete time steps.
In the first step, the neurons in the \emph{input layer}
are \emph{activated} according to the \emph{input vector}.
Then, we iteratively compute output of each neuron in the next layer
until the output layer is reached.
The activity of output layer is then presented as the result.

The activation $y_i$ of neuron $i$ from the layer $I$ is computed as
\begin{equation}
y_i = f\left(\sum_{j \in J}{w_{ij} y_j}\right)
\end{equation}
where $J$ is the previous layer, while $y_j$ is the activation for neurons from $J$ layer.
Function $f()$ is a~so-called \emph{activation function}
and its purpose is to bound the outputs of neurons.
A typical example of an activation function is the sigmoid function.%
\footnote{A special case of the logistic function $\sigma(x)=(1+e^{-(rx+k)})^{-1}$.
Parameters control the growth rate $r$ and the x-position $k$.}

\subsubsection{Training}
Training of the feed-forward neural network usually involves some
modification of supervised Backpropagation learning algorithm.
We use first-order optimization algorithm called RPROP \cite{Riedmiller1993}.
%
%
As outlined above, the training set $T$ consists of
$(\vec p_i, \vec o_i)$ pairs.
The training algorithm is shown in Algorithm \ref{alg:tnn}.

\begin{algorithm}
\caption{Training Neural Network}
\begin{algorithmic}
\label{alg:tnn}
\REQUIRE{Train set $T$, desired error $e$, max iterations $M$}
\STATE $N \leftarrow \mathit{RandomlyInitializedNetwork}()$
\STATE $\mathit{It} \leftarrow 0$
\REPEAT
\STATE $\mathit{It} \leftarrow \mathit{It} + 1$
\STATE $\Delta \vec w \leftarrow \vec 0$
\STATE $\mathit{TotalError} \leftarrow 0$
\FORALL{$(\mathit{Input}, \mathit{DesiredOutput}) \in T$}
\STATE $\mathit{Output} \leftarrow \mathit{Result}(N, \mathit{Input})$
\STATE $\mathit{Error} \leftarrow |\mathit{DesiredOutput} - \mathit{Output}|$
\STATE $\Delta \vec w \leftarrow \Delta \vec w + \mathit{WeightUpdate}(N,\mathit{Error})$
\STATE $\mathit{TotalError} \leftarrow \mathit{TotalError} + \mathit{Error}$
\ENDFOR
\STATE $N \leftarrow \mathit{ModifyWeights}(N, \Delta \vec w)$
\UNTIL{$\mathit{TotalError} < e$ or $ \mathit{It} > M$}
\end{algorithmic}
\end{algorithm}

\subsection{Naive Bayes Classifier}
\label{naive-bayes}

The Naive Bayes Classifier uses existing information to construct
probability model of likelihoods of given {\em feature variables}
based on a discrete-valued {\em class variable}.
Using the Bayes equation, we can then estimate the probability distribution
of class variable for particular values of the feature variables.

In order to approximate the player's output vector $\vec O$ based on
pattern vector $\vec P$, we will compute each element of the
output vector separately, covering the output domain by several $k$-sized
discrete intervals (classes).
\rv{In fact, we use the PCA-represented input $\vec R$ (using the 10 most significant
dimensions), since it better fits the pre-requisites of the
Bayes classifier -- values in each dimension are more independent and
they approximate the normal distribution better. Additionally, small input dimensions
are computationaly feasible.}

When training the classifier for $\vec O$ element $o_i$
of class $c = \lfloor o_i/k \rfloor$,
we assume the $\vec R$ elements are normally distributed and
feed the classifier information in the form
$$ \vec R \mid c $$
estimating the mean $\mu_c$ and standard deviation $\sigma_c$
of each $\vec R$ element for each encountered $c$
(see algorithm \ref{alg:tnb}).
Then, we can query the built probability model on
$$ \max_c P(c \mid \vec R) $$
obtaining the most probable class $i$ for an arbitrary $\vec R$
Each probability is obtained using the normal distribution formula:
$$ P(c \mid x) = {1\over \sqrt{2\pi\sigma_c^2}}\exp{-(x-\mu_c)^2\over2\sigma_c^2} $$

\begin{algorithm}
\caption{Training Naive Bayes}
\begin{algorithmic}
\label{alg:tnb}
\REQUIRE{Training set $T = (\mathit{R, c})$}
\FORALL{$(R, c) \in T$}
\STATE $\mathit{RbyC}_c \leftarrow \mathit{RbyC}_c \cup \{R\}$
\ENDFOR
\FORALL{$c$}
\STATE $\mu_c \leftarrow {1 \over |\mathit{RbyC}_c|} \sum_{R \in \mathit{RbyC}_c} R$
\ENDFOR
\FORALL{$c$}
\STATE $\sigma_c \leftarrow {1 \over |\mathit{RbyC}_c|} \sum_{R \in \mathit{RbyC}_c} R-\mu_c $
\ENDFOR
\end{algorithmic}
\end{algorithm}

\subsection{Implementation}

We have implemented the data mining methods as the
``gostyle'' open-source framework \cite{GoStyle},
made available under the GNU GPL licence.
The majority of our basic processing and \rv{analysis
is} implemented in the Python \cite{Python25} programming language.

We use several external libraries, most notably the MDP library \cite{MDP} \rv{for the PCA analysis}.
The neural network \rv{component} is written using the libfann C library \cite{Nissen2003}.
The Naive Bayes Classifier \rv{is built around} the {\tt AI::NaiveBayes1} Perl module \cite{NaiveBayes1}.
The sociomap has been visualised using the Team Profile Analyzer \cite{TPA}
which is a part of the Sociomap suite \cite{SociomapSite}.

\section{Strength Analysis}
\label{strength-analysis}

\begin{figure*}[!t]
\centering
\includegraphics[width=7in]{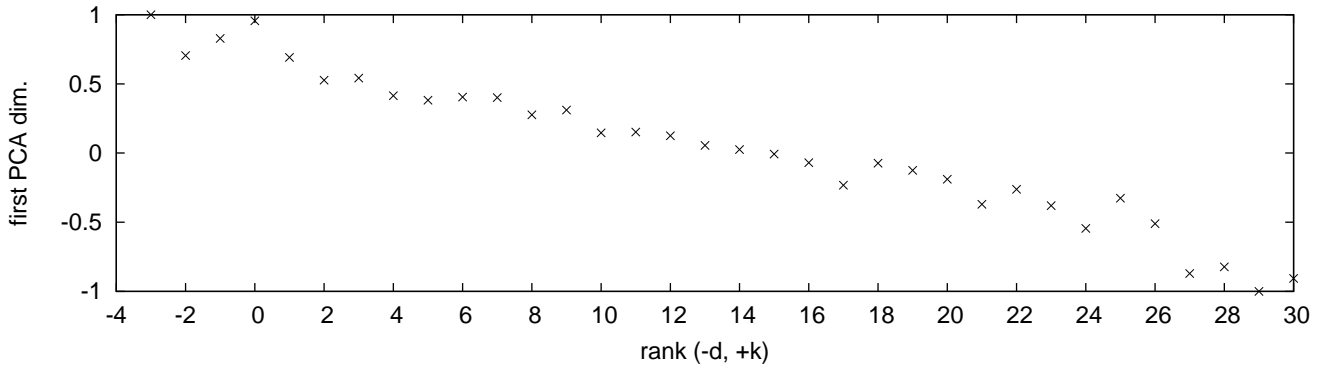}
\caption{PCA of by-strength vectors}
\label{fig:strength_pca}
\end{figure*}

First, we have used our framework to analyse correlations of pattern vectors
and playing strength. Like in other competitively played board games, Go players
receive real-world {\em rating number} based on tournament games,
and {\em rank} based on their rating.
The amateur ranks range from 30-kyu (beginner) to 1-kyu (intermediate)
and then follows 1-dan to 9-dan
(top-level player).

There are multiple independent real-world ranking scales
(geographically based), \rv{while} online servers \rv{also} maintain their own user rank \rv{list}\rvvv{.
The} difference between scales can be up to several ranks and the rank
distributions also differ \cite{RankComparison}.

\subsection{Data source}
As the source game collection, we use \rvv{the} Go Teaching Ladder reviews archive
\cite{GTL}. This collection contains 7700 games of players with strength ranging
from 30-kyu to 4-dan; we consider only even games with clear rank information.

Since the rank information is provided by the users and may not be consistent,
we are forced to take a simplified look at the ranks,
discarding the differences between various systems and thus somewhat
increasing error in our model.\footnote{Since our results seem satisfying,
we did not pursue to try another collection;
one could e.g. look at game archives of some Go server to work within
single more-or-less consistent rank model.}
We represent the rank in our dataset \rv{as an integer in the range} $[-3,30]$ with positive
numbers representing the kyu ranks and numbers smaller than 1 representing the dan
ranks: 4-dan maps to $-3$, 1-dan to $0$, etc.

\subsection{Strength PCA analysis}
First, we have created a single pattern vector for each rank between 30-kyu to 4-dan;
we have performed PCA analysis on the pattern vectors, achieving near-perfect
rank correspondence in the first PCA dimension%
\footnote{The eigenvalue of the second dimension was four times smaller,
with no discernable structure revealed within the lower-order eigenvectors.}
(figure \ref{fig:strength_pca}).
We measure the accuracy of the strength approximation by the first PCA dimension
using Pearson's $r$ (see \ref{pearson}), yielding very satisfying value of $r=0.979$
implying extremely strong correlation.%

\rv{This reflects the trivial fact that the most important ``defining characteristic''
of a set of players grouped by strength is indeed their strength and confirms
that our methodics is correct.
At the same time, this result suggests that it is possible to accurately estimate
player's strength \rvvv{just} from a sample of his games, 
as we confirm below.}

\rv{When investigating a player's $\vec p$, the PCA decomposition could be also
useful for study suggestions --- a program could examine the pattern gradient at the
player's position on the PCA dimensions and suggest patterns to avoid and patterns
to play more often. Of course, such an advice alone is certainly not enough and it
must be used only as a basis of a more thorough analysis of reasons behind the fact
that the player plays other patterns than they ``should''.}

\subsection{Strength Classification}
\label{strength-class}

\rv{In line with results of the PCA analysis, we have tested the strength approximation ability
of $k$-NN (sec.} \ref{knn}\rv{), neural network (sec. }\ref{neural-net}\rv{),
and a simple PCA-based classifier (sec. }\ref{PCA}\rv{).}

\subsubsection{Reference (Training) Data}
\rv{We have trained the tested classifiers using one pattern vector per rank
(aggregate over all games played by some player declaring the given rank),
then performing PCA analysis to reduce the dimension of pattern vectors.}
We have explored the influence of different game sample sizes (\rv{$G$})
on the classification accuracy to \rv{determine the} practicality and scaling
abilities of the classifiers.
In order to reduce the diversity of patterns (negatively impacting accuracy
on small samples), we do not consider the contiguity pattern features.

The classifiers were compared by running a many-fold validation by repeatedly and
exhaustively taking disjunct \rv{$G$}--game samples of the same rank from the collection
and measuring the standard error of the classifier.
Arbitrary game numbers were approximated by pattern file sizes,
iteratively selecting all games of randomly selected player
of the required strength.

%

\subsubsection{Results}
\rv{The results are shown in the table~}\ref{table-str-class}\rv{.
The $G$ column describes the number of games in each sample,
$\mathit{MSE}$ column shows measured mean square error and $\sigma$ is the empirical standard deviation.
Methods are
compared (column $\mathit{Cmp}$) to the random classifier by the quotient of their~$\sigma$.}

\rv{From the table, it should be obvious that the $k$-NN is obtaining good
accuracy even on as few as 9 games as a sample\rvv{, where the classifier performs within a standard deviation of $4.6$kyu.}
For a large number of training vectors -- albeit not very accurate due to small
sample sizes -- the neural network classifier performs very similarly.
For samples of 2 games, the neural network is even slightly better on average.
However, due to the decreasing number of training vectors with increasing game sample sizes,
the neural network gets unusable for large \rvvvv{game} sample sizes.
The table therefore only shows the neural network results for samples of 17 games and smaller.}
\rv{PCA-based classifier (the most significant PCA eigenvector position is simply directly taken as a~rank) and
a random classifier are listed mainly for the sake of comparison, because they do not perform
competetively.}

\begin{table}[!t]
\renewcommand{\arraystretch}{1.3}
\caption{Strength Classifier Performance}
\label{table-str-class}
\centering
\begin{tabular}{|c|c||c|c||c|}
\hline
Method & \rv{$G$} & MSE & $\sigma$ & Cmp \\ \hline
$k$-NN&$85$ & $5.514$ & $2.348$ & $6.150$ \\
&$43$ & $8.449$ & $2.907$ & $4.968$ \\
&$17$ & $10.096$& $3.177$ & $4.545$ \\
&$9$  & $21.343$& $4.620$ & $3.126$ \\
&$2$  & $52.212$& $7.226$ & $1.998$ \\\hline

\rv{Neural Network} & $17$ & $110.633$ & $10.518$ & $1.373$ \\
&$9$  & $44.512$ & $6.672$ & $2.164$ \\
&$2$  & $43.682$ & $6.609$ & $2.185$ \\ \hline

PCA & $85$ & $24.070$ & $4.906$ & $2.944$ \\
&$43$ & $31.324$ & $5.597$ & $2.580$ \\
&$17$ & $50.390$ & $7.099$ & $2.034$ \\
&$9$  & $72.528$ & $8.516$ & $1.696$ \\
&$2$  & $128.660$& $11.343$ & $1.273$ \\ \hline

Rnd & N/A & $208.549$ & $14.441$ & $1.000$ \\ \hline
\end{tabular}
\end{table}

\subsubsection{$k$-NN parameters}
\rv{Using the $4$-Nearest Neighbors classifier with the weight function}
\begin{equation}
\mathit{Weight}(\vec x) = 0.9^{M*\mathit{Distance}(\vec x)}
\end{equation}
(parameter $M$ ranging from $30$ to $6$).

\subsubsection{Neural network's parameters}
\rv{The neural network classifier had three-layered architecture (one hidden layer)
comprising of these numbers of neurons:}
\vspace{4mm}
\noindent
\begin{center}
\begin{tabular}{|c|c|c|}
\hline
\multicolumn{3}{|c|}{Layer} \\\hline
Input & Hidden & Output \\ \hline
 119 & 35 & 1 \\ \hline
\end{tabular}
\end{center}
\vspace{4mm}

\rv{The network was trained until the square error on the training set was smaller than $0.0005$.
Due to the small number of input vectors,
this only took about $20$ iterations of RPROP learning algorithm on average.}


\section{Style Analysis}
\label{style-analysis}

As a~second case study for our pattern analysis,
we investigate pattern vectors $\vec p$ of various well-known players,
their relationships in-between and to prior knowledge
in order to explore the correlation of prior knowledge with extracted patterns.
We look for relationships between pattern vectors and perceived
``playing style'' and attempt to use our classifiers to transform
the pattern vector $\vec p$ to a style vector $\vec s$.

\subsection{Data sources}
\subsubsection{Game database}
The source game collection is GoGoD Winter 2008 \cite{GoGoD} containing 55000
professional games, dating from the early Go history 1500 years ago to the present.
We consider only games of a small subset of players (table \ref{fig:style_marks})\rvvv{. These players
were chosen} for being well-known within the players community,
having large number of played games in our collection and not playing too long
ago.

\subsubsection{Expert-based knowledge}
\label{style-vectors}
In order to provide a reference frame for our style analysis,
we have gathered some information from game experts about various
traditionally perceived style aspects to use as a prior knowledge.
This expert-based knowledge allows us to predict styles of unknown players
based on the similarity of their pattern vectors,
as well as discover correlations between styles and \rv{particular}
move patterns.

Experts were asked to mark four style aspects of each of the given players
on the scale from 1 to 10. The style aspects are defined as shown:

\vspace{4mm}
\noindent
\begin{center}
\begin{tabular}{|c|c|c|}
\hline
Style & 1 & 10\\ \hline
Territoriality $\tau$ & Moyo & Territory \\
Orthodoxity $\omega$ & Classic & Novel \\
Aggressivity $\alpha$ & Calm & Fighting \\
Thickness $\theta$ & Safe & Shinogi \\ \hline
\end{tabular}
\end{center}
\vspace{4mm}

We have devised these four style aspects based on our own Go experience
and consultations with other experts.
The used terminology has quite
clear meaning to any experienced Go player and there is not too much
room for confusion, except possibly in the case of ``thickness'' ---
but the concept is not easy to pin-point succintly and we also did not
add extra comments on the style aspects to the questionnaire deliberately
to accurately reflect any diversity in understanding of the terms.
Averaging this expert based evaluation yields \emph{reference style vector}
$\vec s_r$ (of dimension $4$) for each player $r$
from the set of \emph{reference players} $R$.

Throughout our research, we have experimentally found that playing era
is also a major factor differentiating between patterns. Thus, we have
further extended the $\vec s_r$ by median year over all games played
by the player.

\begin{table}[!t]
\renewcommand{\arraystretch}{1.3}
\caption{Covariance Measure of Prior Information Dimensions}
\label{fig:style_marks_r}
\centering
\begin{tabular}{|r||r||r||r||r||r|}
\hline
  & $\tau$ & $\omega$ & $\alpha$ & $\theta$ & year \\
\hline
$\tau$  &$1.000$&$\mathbf{-0.438}$&$\mathbf{-0.581}$&$\mathbf{ 0.721}$&$ 0.108$\\
$\omega$&       &$ 1.000$&$\mathbf{ 0.682}$&$ 0.014$&$-0.021$\\
$\alpha$&       &        &$ 1.000$&$-0.081$&$ 0.030$\\
$\theta$&       &\multicolumn{1}{c||}{---}
                         &        &$ 1.000$&$-0.073$\\
y.      &       &        &        &        &$ 1.000$\\
\hline
\end{tabular}
\end{table}

Three high-level Go players (Alexander Dinerstein 3-pro, Motoki Noguchi
7-dan and V\'{i}t Brunner 4-dan) have judged the style of the reference
players.
The complete list of answers is in table \ref{fig:style_marks}.
Standard error of the answers is 0.952, making the data reasonably reliable,
though much larger sample would of course be more desirable
(but beyond our means to collect).
We have also found a~significant correlation between the various
style aspects, as shown by the Pearson's $r$ values
in table \ref{fig:style_marks_r}.

\rv{We have made few manual adjustments in the dataset, disregarding some
players or portions of their games. This was done to achieve better
consistency of the games (e.g. considering only games of roughly the
same age) and to consider only sets of games that can be reasonably
rated as a whole by human experts (who can give a clear feedback in this
effect). This filtering methodology can be easily reproduced
and such arbitrary decisions are neccessary only
for processing the training dataset, not for using it (either for exloration
or classification).}

\begin{table}[!t]
\renewcommand{\arraystretch}{1.4}
\begin{threeparttable}
\caption{Expert-Based Style Aspects of Selected Professionals\tnote{1} \tnote{2}}
\label{fig:style_marks}
\centering
\begin{tabular}{|c||c||c||c||c|}
\hline
{Player} & $\tau$ & $\omega$ & $\alpha$ & $\theta$ \\
\hline
Go Seigen\tnote{3}   & $6.0 \pm 2.0$ & $9.0 \pm 1.0$ & $8.0 \pm 1.0$ & $5.0 \pm 1.0$ \\
Ishida Yoshio\tnote{4}&$8.0 \pm 1.4$ & $5.0 \pm 1.4$ & $3.3 \pm 1.2$ & $5.3 \pm 0.5$ \\
Miyazawa Goro        & $1.5 \pm 0.5$ & $10  \pm 0  $ & $9.5 \pm 0.5$ & $4.0 \pm 1.0$ \\
Yi Ch'ang-ho\tnote{5}& $7.0 \pm 0.8$ & $5.0 \pm 1.4$ & $2.6 \pm 0.9$ & $2.6 \pm 1.2$ \\
Sakata Eio           & $7.6 \pm 1.7$ & $4.6 \pm 0.5$ & $7.3 \pm 0.9$ & $8.0 \pm 1.6$ \\
Fujisawa Hideyuki    & $3.5 \pm 0.5$ & $9.0 \pm 1.0$ & $7.0 \pm 0.0$ & $4.0 \pm 0.0$ \\
Otake Hideo          & $4.3 \pm 0.5$ & $3.0 \pm 0.0$ & $4.6 \pm 1.2$ & $3.6 \pm 0.9$ \\
Kato Masao           & $2.5 \pm 0.5$ & $4.5 \pm 1.5$ & $9.5 \pm 0.5$ & $4.0 \pm 0.0$ \\
Takemiya Masaki\tnote{4}&$1.3\pm 0.5$& $6.3 \pm 2.1$ & $7.0 \pm 0.8$ & $1.3 \pm 0.5$ \\
Kobayashi Koichi     & $9.0 \pm 1.0$ & $2.5 \pm 0.5$ & $2.5 \pm 0.5$ & $5.5 \pm 0.5$ \\
Cho Chikun           & $9.0 \pm 0.8$ & $7.6 \pm 0.9$ & $6.6 \pm 1.2$ & $9.0 \pm 0.8$ \\
Ma Xiaochun          & $8.0 \pm 2.2$ & $6.3 \pm 0.5$ & $5.6 \pm 1.9$ & $8.0 \pm 0.8$ \\
Yoda Norimoto        & $6.3 \pm 1.7$ & $4.3 \pm 2.1$ & $4.3 \pm 2.1$ & $3.3 \pm 1.2$ \\
Luo Xihe             & $7.3 \pm 0.9$ & $7.3 \pm 2.5$ & $7.6 \pm 0.9$ & $6.0 \pm 1.4$ \\
O Meien              & $2.6 \pm 1.2$ & $9.6 \pm 0.5$ & $8.3 \pm 1.7$ & $3.6 \pm 1.2$ \\
Rui Naiwei           & $4.6 \pm 1.2$ & $5.6 \pm 0.5$ & $9.0 \pm 0.8$ & $3.3 \pm 1.2$ \\
Yuki Satoshi         & $3.0 \pm 1.0$ & $8.5 \pm 0.5$ & $9.0 \pm 1.0$ & $4.5 \pm 0.5$ \\
Hane Naoki           & $7.5 \pm 0.5$ & $2.5 \pm 0.5$ & $4.0 \pm 0.0$ & $4.5 \pm 1.5$ \\
Takao Shinji         & $5.0 \pm 1.0$ & $3.5 \pm 0.5$ & $5.5 \pm 1.5$ & $4.5 \pm 0.5$ \\
Yi Se-tol            & $5.3 \pm 0.5$ & $6.6 \pm 2.5$ & $9.3 \pm 0.5$ & $6.6 \pm 1.2$ \\
Yamashita Keigo\tnote{4}&$2.0\pm 0.0$& $9.0 \pm 1.0$ & $9.5 \pm 0.5$ & $3.0 \pm 1.0$ \\
Cho U                & $7.3 \pm 2.4$ & $6.0 \pm 0.8$ & $5.3 \pm 1.7$ & $6.3 \pm 1.7$ \\
Gu Li                & $5.6 \pm 0.9$ & $7.0 \pm 0.8$ & $9.0 \pm 0.8$ & $4.0 \pm 0.8$ \\
Chen Yaoye           & $6.0 \pm 1.0$ & $4.0 \pm 1.0$ & $6.0 \pm 1.0$ & $5.5 \pm 0.5$ \\
\hline
\end{tabular}
\begin{tablenotes}
\item [1] Including standard deviation. Only players where we received at least two out of three answers are included.
\item [2] Since the playing era column does not fit into the table, we at least sort the players ascending by their median year.
\item [3] We do not consider games of Go Seigen due to him playing across several distinct eras and also being famous for radical opening experiments throughout the time, and thus featuring especially high diversity in patterns.
\item [4] We do not consider games of Ishida Yoshio and Yamashita Keigo for the PCA analysis since they are significant outliers, making high-order dimensions much like purely ``similarity to this player''. Takemiya Masaki has the similar effect for the first dimension, but that case corresponds to common knowledge of him being an extreme proponent of anti-territorial (``moyo'') style.
\item [5] We consider games only up to year 2004, since Yi Ch'ang-ho was prominent representative of a balanced, careful player until then and still has this reputation in minds of many players, but is regarded to have altered his style significantly afterwards.
\end{tablenotes}
\end{threeparttable}
\end{table}

\subsection{Style PCA analysis}

\begin{figure}[!t]
\centering
\includegraphics[width=3in]{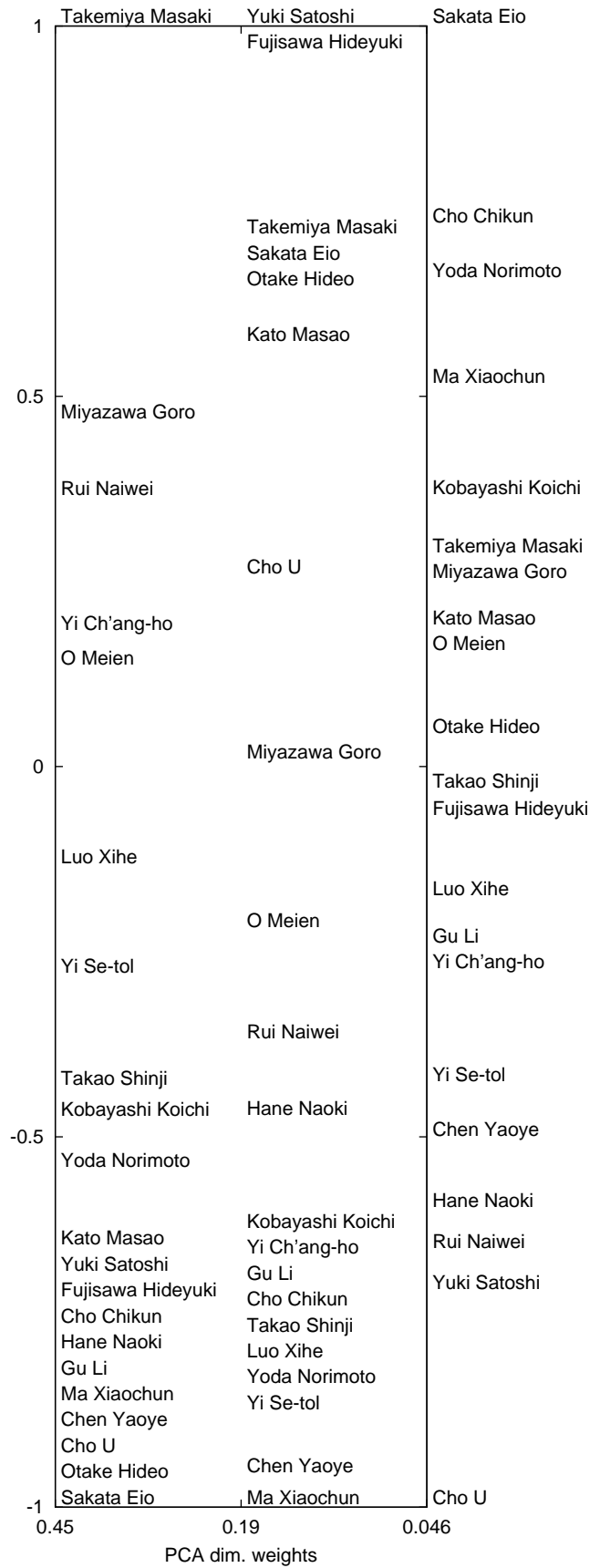}
\caption{Columns with the most significant PCA dimensions of the dataset.}
\label{fig:style_pca}
\end{figure}

We have looked at the ten most significant dimensions of the pattern data
yielded by the PCA analysis of the reference player set
(fig. \ref{fig:style_pca} shows the first three).
We have again computed the Pearson's $r$ for all combinations of PCA dimensions
and dimensions of the prior knowledge style vectors to find correlations.

\begin{table}[!t]
\renewcommand{\arraystretch}{1.4}
\caption{Covariance Measure of PCA and Prior Information}
\label{fig:style_r}
\centering
\begin{tabular}{|c||r||r||r||r||r|}
\hline
Eigenval. & $\tau$ & $\omega$ & $\alpha$ & $\theta$ & Year \\
\hline
$0.447$ & $\mathbf{-0.530}$ & $ 0.323$ & $ 0.298$ & $\mathbf{-0.554}$ & $ 0.090$ \\
$0.194$ & $\mathbf{-0.547}$ & $ 0.215$ & $ 0.249$ & $-0.293$ & $\mathbf{-0.630}$ \\
$0.046$ & $ 0.131$ & $-0.002$ & $-0.128$ & $ 0.242$ & $\mathbf{-0.630}$ \\
$0.028$ & $-0.011$ & $ 0.225$ & $ 0.186$ & $ 0.131$ & $ 0.067$ \\
$0.024$ & $-0.181$ & $ 0.174$ & $-0.032$ & $-0.216$ & $ 0.352$ \\
\hline
\end{tabular}
\end{table}

\begin{table}[!t]
\renewcommand{\arraystretch}{1.6}
\begin{threeparttable}
\caption{Characteristic Patterns of PCA$_{1,2}$ Dimensions \tnote{1}}
\label{fig:style_patterns}
\centering
\begin{tabular}{|p{2.3cm}p{2.4cm}p{2.4cm}p{0cm}|}

\hline \multicolumn{4}{|c|}{PCA$_1$ --- Moyo-oriented, thin-playing player} \\
\centering \begin{psgopartialboard*}{(8,1)(12,6)}
\stone[\marktr]{black}{k}{4}
\end{psgopartialboard*} &
\centering \begin{psgopartialboard*}{(1,2)(5,6)}
\stone{white}{d}{3}
\stone[\marktr]{black}{d}{5}
\end{psgopartialboard*} &
\centering \begin{psgopartialboard*}{(5,1)(10,6)}
\stone{white}{f}{3}
\stone[\marktr]{black}{j}{4}
\end{psgopartialboard*} & \\
\centering $0.274$ & \centering $0.086$ & \centering $0.083$ & \\
\centering high corner/side opening \tnote{2} & \centering high corner approach & \centering high distant pincer & \\

\hline \multicolumn{4}{|c|}{PCA$_1$ --- Territorial, thick-playing player} \\
\centering \begin{psgopartialboard*}{(3,1)(7,6)}
\stone{white}{d}{4}
\stone[\marktr]{black}{f}{3}
\end{psgopartialboard*} &
\centering \begin{psgopartialboard*}{(3,1)(7,6)}
\stone{white}{c}{6}
\stone{black}{d}{4}
\stone[\marktr]{black}{f}{3}
\end{psgopartialboard*} &
\centering \begin{psgopartialboard*}{(3,1)(7,6)}
\stone{black}{d}{4}
\stone[\marktr]{black}{f}{3}
\end{psgopartialboard*} & \\
\centering $-0.399$ & \centering $-0.399$ & \centering $-0.177$ & \\
\centering low corner approach & \centering low corner reply & \centering low corner enclosure & \\

\hline \multicolumn{4}{|c|}{PCA$_2$ --- Territorial, current player \tnote{3}} \\
\centering \begin{psgopartialboard*}{(3,1)(7,6)}
\stone{white}{c}{6}
\stone{black}{d}{4}
\stone[\marktr]{black}{f}{3}
\end{psgopartialboard*} &
\centering \begin{psgopartialboard*}{(3,1)(8,6)}
\stone{white}{d}{4}
\stone[\marktr]{black}{g}{4}
\end{psgopartialboard*} &
\centering \begin{psgopartialboard*}{(4,1)(9,6)}
\stone{black}{d}{4}
\stone{white}{f}{3}
\stone[\marktr]{black}{h}{3}
\end{psgopartialboard*} & \\
\centering $-0.193$ & \centering $-0.139$ & \centering $-0.135$ & \\
\centering low corner reply \tnote{4} & \centering high distant approach/pincer & \centering near low pincer & \\

\hline
\end{tabular}
\begin{tablenotes}
\item [1] We present the patterns in a simplified compact form;
in reality, they are usually somewhat larger and always circle-shaped
(centered on the triangled move).
We omit only pattern segments that are entirely empty.
\item [2] We give some textual interpretation of the patterns, especially
since some of them may not be obvious unless seen in game context; we choose
the descriptions based on the most frequently observer contexts, but of course
the pattern can be also matched in other positions and situations.
\item [3] In the second PCA dimension, we find no correlated patterns;
only uncorrelated and anti-correlated ones.
\item [4] As the second most significant pattern,
we skip a slide follow-up pattern to this move.
\end{tablenotes}
\end{threeparttable}
\end{table}

\begin{table}[!t]
\renewcommand{\arraystretch}{1.8}
\begin{threeparttable}
\caption{Characteristic Patterns of PCA$_3$ Dimension \tnote{1}}
\label{fig:style_patterns3}
\centering
\begin{tabular}{|p{2.4cm}p{2.4cm}p{2.4cm}p{0cm}|}

\hline \multicolumn{4}{|c|}{PCA$_3$ --- Old-time player} \\
\centering \begin{psgopartialboard*}{(1,3)(5,7)}
\stone{white}{d}{4}
\stone[\marktr]{black}{c}{6}
\end{psgopartialboard*} &
\centering \begin{psgopartialboard*}{(8,1)(12,5)}
\stone[\marktr]{black}{k}{3}
\end{psgopartialboard*} &
\centering \begin{psgopartialboard*}{(1,1)(5,5)}
\stone[\marktr]{black}{c}{3}
\end{psgopartialboard*} & \\
\centering $0.515$ & \centering $0.264$ & \centering $0.258$ & \\
\centering low corner approach & \centering low side or mokuhazushi opening & \centering san-san opening & \\

\hline \multicolumn{4}{|c|}{PCA$_3$ --- Current player} \\
\centering \begin{psgopartialboard*}{(3,1)(7,5)}
\stone{black}{d}{4}
\stone[\marktr]{black}{f}{3}
\end{psgopartialboard*} &
\centering \begin{psgopartialboard*}{(1,1)(5,5)}
\stone[\marktr]{black}{c}{4}
\end{psgopartialboard*} &
\centering \begin{psgopartialboard*}{(1,2)(5,6)}
\stone{black}{d}{3}
\stone{white}{d}{5}
\stone[\marktr]{black}{c}{5}
\end{psgopartialboard*} & \\
\centering $-0.276$ & \centering $-0.273$ & \centering $-0.116$ & \\
\centering low corner enclosure & \centering 3-4 corner opening \tnote{2} & \centering high approach reply & \\

\hline
\end{tabular}
\begin{tablenotes}
\item [1] We cannot use terms ``classic'' and ''modern'' in case of PCA$_3$
since the current patterns are commonplace in games of past centuries
(not included in our training set) and many would call a lot of the old-time patterns
modern inventions. Perhaps we can infer that the latest 21th-century play trends abandon
many of the 20th-century experiments (lower echelon of our by-year samples)
to return to the more ordinary but effective classic patterns.
\item [2] At this point, we skip two patterns already shown elsewhere:
{\em high side/corner opening} and {\em low corner reply}.
\end{tablenotes}
\end{threeparttable}
\end{table}

It is immediately
obvious both from the measured $r$ and visual observation
that by far the most significant vector corresponds very well
to the territoriality of the players,
confirming the intuitive notion that this aspect of style
is the one easiest to pin-point and also
most obvious in the played shapes and sequences
(that can obviously aim directly at taking secure territory
or building center-oriented framework). Thick (solid) play also plays
a role, but these two style dimensions are already
correlated in the prior data.

The other PCA dimensions are somewhat harder to interpret, but there
certainly is significant influence of the styles on the patterns;
the correlations are all presented in table \ref{fig:style_r}.
(Larger absolute value means better linear correspondence.)

We also list the characteristic spatial patterns of the PCA dimension
extremes (tables \ref{fig:style_patterns}, \ref{fig:style_patterns3}), determined by their coefficients
in the PCA projection matrix --- however, such naive approach
has limited reliability, better methods will have to be researched.%
\footnote{For example, as one of highly ranked ``Takemiya's'' PCA1 patterns,
3,3 corner opening was generated, completely inappropriately;
it reflects some weak ordering in bottom half of the dimension,
not global ordering within the dimension.}
We do not show the other pattern features since they carry no useful
information in the opening stage.
\begin{table}[!t]
\renewcommand{\arraystretch}{1.4}
\caption{Covariance Measure of Externed-Normalization PCA and~Prior Information}
\label{fig:style_normr}
\centering
\begin{tabular}{|c||r||r||r||r||r|}
\hline
Eigenval. & $\tau$ & $\omega$ & $\alpha$ & $\theta$ & Year \\
\hline
$6.377$ & $ \mathbf{0.436}$ & $-0.220$ & $-0.289$ & $ \mathbf{0.404}$ & $\mathbf{-0.576}$ \\
$1.727$ & $\mathbf{-0.690}$ & $ 0.340$ & $ 0.315$ & $\mathbf{-0.445}$ & $\mathbf{-0.639}$ \\
$1.175$ & $-0.185$ & $ 0.156$ & $ 0.107$ & $-0.315$ & $ 0.320$ \\
$0.845$ & $ 0.064$ & $-0.102$ & $-0.189$ & $ 0.032$ & $ 0.182$ \\
$0.804$ & $-0.185$ & $ 0.261$ & $ \mathbf{0.620}$ & $ 0.120$ & $ 0.056$ \\
$0.668$ & $-0.027$ & $ 0.055$ & $ 0.147$ & $-0.198$ & $ 0.155$ \\
$0.579$ & $ 0.079$ & $ \mathbf{0.509}$ & $ 0.167$ & $ 0.294$ & $-0.019$ \\
\hline
\end{tabular}
\end{table}

The PCA results presented above do not show much correlation between
the significant PCA dimensions and the $\omega$ and $\alpha$ style dimensions.
However, when we applied the extended vector normalization
(sec. \ref{xnorm}; see table \ref{fig:style_normr}),
some less significant PCA dimensions exhibited clear correlations.%
\footnote{We have found that $c=6$ in the post-processing logistic function
produces the most instructive PCA output on our particular game collection.}
While we do not use the extended normalization results elsewhere since
they produced noticeably less accurate classifiers in all dimensions
(including $\omega$ and $\alpha$), it is instructive to look at the PCA dimensions.

\rv{In contrast with the emphasis of opening patterns in the $\tau$ and $\theta$
dimensions, the most contributing patterns of the $\omega$ and $\alpha$
dimensions are the middle-game patterns that occur less frequently and require
the extended normalization not to be over-shadowed by the opening patterns.}%
\footnote{In the middle game, \rv{basic areas of influence have been staked
out and invasions and group attacks are being played out}.
Notably, the board is much more filled \rv{than in the opening} and thus
particular specific-shape patterns repeat less often.}
E.g. the most characteristic patterns
on the aggressiveness dimension represent moves that make life with small,
unstable groups (connecting kosumi on second line or mouth-shape eyespace
move), while the novel-ranked players seem to like the (in)famous tsuke-nobi
joseki sequence.%
\footnote{\rv{Tsuke-nobi is a well-known joseki popular among beginners,
but professionals usually play it only in special contexts.}}
\rv{This may either mean that novel players like to play the joseki more,
or (more likely, in our opinion) that novel players are more likely to
get into unorthodox situation that require resorting to the tsuke-nobi
sequence.}
We believe that the next step in interpreting our analytical results
will be more refined prior information input
and precise analysis of the outputs by Go experts.

\begin{figure}[!t]
\centering
\includegraphics[width=3.5in,angle=-90]{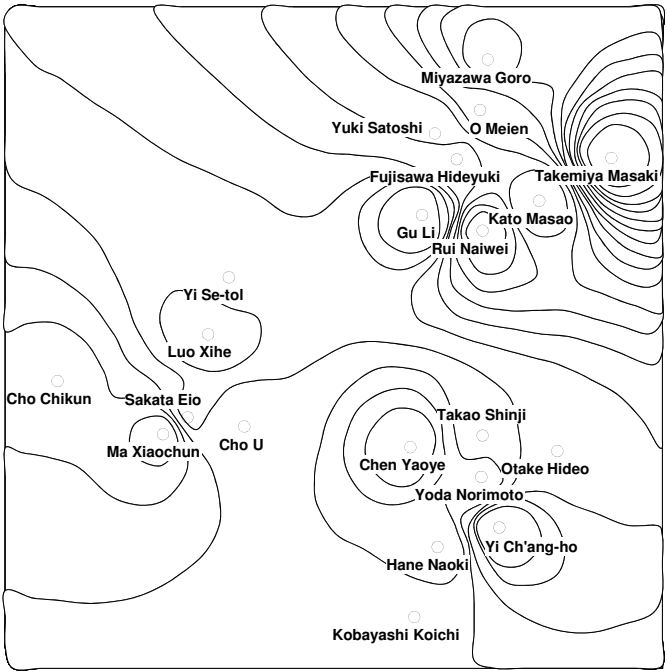}
\caption{Sociomap visualisation. The spatial positioning of players
is based on the expert knowledge, while the node heights (depicted by
contour lines) represent the pattern vectors.%
}
\label{fig:sociomap}
\end{figure}

Fig. \ref{fig:sociomap} shows the Sociomap visualisation
as an alternate view of the player relationships and similarity,
as well as correlation between the expert-given style marks
and the PCA decomposition. The four-dimensional style vectors
are used as input for the Sociomap renderer and determine the
spatial positions of players. The height of a node is then
determined using first two PCA dimensions $R_1,R_2$ and their
eigenvalues $\lambda_1,\lambda_2$ as their linear combination:
$$ h=\lambda_1R_1 + \lambda_2R_2 $$

We can observe that the terrain of the sociomap is reasonably
``smooth'', again demonstrating some level of connection between
the style vectors and data-mined information. High countour density
indicates some discrepancy; in case of Takemiya Masaki and Yi Ch'ang-ho,
this seems to be merely an issue of scale,
while the Rui Naiwei --- Gu Li cliff suggests a genuine problem;
we cannot say now whether it is because of imprecise prior information
or lacking approximation abilities of our model.

\subsection{Style Classification}
\label{style-class}


Similarly to the the Strength classification (section \ref{strength-class}), we have tested the style inference ability
of $k$-NN (sec. \ref{knn}), neural network (sec. \ref{neural-net}), and Bayes (sec. \ref{naive-bayes}) classifers.

\subsubsection{Reference (Training) Data}
As the~reference data, we use expert-based knowledge presented in section \ref{style-vectors}.
For each reference player, that gives $4$-dimensional \emph{style vector} (each component in the
range of $[1,10]$).\footnote{Since the neural network has activation function with range $[-1,1]$, we
have linearly rescaled the \emph{style vectors} from interval $[1,10]$ to $[-1,1]$ before using the training
data. The network's output was afterwards rescaled back to allow for MSE comparison.}

All input (pattern) vectors were preprocessed using PCA, reducing the input dimension from $400$ to $23$.
We measure the performance on the same reference data using $5$-fold cross validation.
To put our measurements in scale, we also include a~random classifier in our results.

\subsubsection{Results}
The results are shown in the table \ref{crossval-cmp}. Second to fifth columns in the table represent
mean square error (MSE) of the examined styles, $\mathit{Mean}$ is the
mean square error across the styles and finally, the last column $\mathit{Cmp}$
represents $\mathit{Mean}(\mathit{Random classifier}) / \mathit{Mean}(\mathit{X})$ -- comparison of mean square error
of each method with the random classifier. To minimize the
effect of random variables, all numbers were taken as an average of $200$ runs of the cross validation.

Analysis of the performance of $k$-NN classifier for different $k$-values has shown that different
$k$-values are suitable to approximate different styles. Combining the $k$-NN classifiers with the 
neural network (so that each style is approximated by the method with lowest MSE in that style)
results in \emph{Joint classifier}, which outperforms all other methods.
The \emph{Joint classifier} has outstanding MSE $3.960$, which is equivalent to standard error
of $\sigma = 1.99$ per style.%
\footnote{We should note that the pattern vector for each tested player
was generated over at least few tens of games.}

\begin{table}[!t]
\renewcommand{\arraystretch}{1.4}
\begin{threeparttable}
\caption{Comparison of style classifiers}
\label{crossval-cmp}
\begin{tabular}{|c|c|c|c|c|c|c|}
\hline
&\multicolumn{5}{|c|}{MSE}& \\ \hline
{Classifier} & $\tau$ & $\omega$ & $\alpha$ & $\theta$ & {\bf Mean} & {\bf Cmp}\\ \hline
Joint classifier\tnote{1} & 4.04 & {\bf 5.25} & {\bf 3.52} & {\bf 3.05} & {\bf 3.960}& 2.97 \\\hline
Neural network   & {\bf 4.03} & 6.15       & {\bf 3.58} & 3.79       & 4.388      & 2.68 \\ 
$k$-NN ($k=2$)   & 4.08       & 5.40       & 4.77       & 3.37       & 4.405      & 2.67 \\
$k$-NN ($k=3$)   & 4.05       & 5.58       & 5.06       & 3.41       & 4.524      & 2.60 \\
$k$-NN ($k=1$)   & 4.52       & {\bf 5.26} & 5.36       & {\bf 3.09} & 4.553      & 2.59 \\
$k$-NN ($k=4$)   & 4.10       & 5.88       & 5.16       & 3.60       & 4.684      & 2.51 \\
Naive Bayes      & 4.48       & 6.90       & 5.48       & 3.70       & 5.143      & 2.29 \\
Random class.    & 12.26      & 12.33      & 12.40      & 10.11      & 11.776     & 1.00 \\\hline

\end{tabular}
\begin{tablenotes}
\item [1] Note that these measurements have a certain variance.
Since the Joint classifier performance was measured from scratch,
the precise values may not match appropriate cells of the used methods.
\end{tablenotes}
\end{threeparttable}
\end{table}

\subsubsection{$k$-NN parameters}
All three variants of $k$-NN classifier ($k=2,3,4$) had the weight function
\begin{equation}
\mathit{Weight}(\vec x) = 0.8^{10*\mathit{Distance}(\vec x)}
\end{equation}
The parameters were chosen empirically to minimize the MSE.

\subsubsection{Neural network's parameters}
The neural network classifier had three-layered architecture (one hidden layer)
comprising of these numbers of neurons:
\vspace{4mm}
\noindent
\begin{center}
\begin{tabular}{|c|c|c|}
\hline
\multicolumn{3}{|c|}{Layer} \\\hline
Input & Hidden & Output \\ \hline
 23 & 30 & 4 \\ \hline
\end{tabular}
\end{center}
\vspace{4mm}

The network was trained until the square error on the training set was smaller than $0.0003$.
Due to a small number of input vectors, this only took $20$ iterations of RPROP learning algorithm on average.

\subsubsection{Naive Bayes parameters}

We have chosen $k = 10/7$ as our discretization parameter;
ideally, we would use $k = 1$ to fully cover the style marks
domain, however our training sample \rv{turns out to be} too small for
that.

\section{Proposed Applications}
\label{proposed-apps-and-discussion}


We believe that our findings might be useful for many applications
in the area of Go support software as well as Go-playing computer engines.
\rv{However, our foremost aim is to use the style analysis as an excellent
teaching aid} --- classifying style
dimensions based on player's pattern vector, many study recommendations
can be given, e.g. about the professional games to replay, the goal being
balancing understanding of various styles to achieve well-rounded skill set.%
\footnote{\rv{The strength analysis could be also used in a similar fashion,
but the lesson learned cannot simply be ``play pattern $X$ more often'';
instead, the insight lays in the underlying reason of disproportionate
frequency of some patterns.}}
\rv{A user-friendly tool based on our work is currently in development.}

\rv{Another promising application we envision is helping to}
determine \rvvvv{the} initial real-world rating of a player before their
first tournament based on a sample of their games played on the internet;
some players especially in less populated areas could get fairly strong
before playing in their first real tournament.
\rv{Similarly, a computer Go program can quickly} classify the level of its
\rv{human opponent} based on the pattern vector from \rv{their previous games}
and auto-adjust its difficulty settings accordingly
to provide more even games for beginners.
\rvvv{This can also be achieved using} win-loss statistics,
but pattern vector analysis \rv{should} converge faster \rv{initially,
providing} \rvvvv{a much} \rv{better user experience}.

We hope that more strong players will look into the style dimensions found
by our statistical analysis --- analysis of most played patterns of prospective
opponents might prepare for \rv{a tournament} game, but we especially hope that new insights
on strategic purposes of various shapes and general human understanding
of the game might be \rv{improved} by investigating the style-specific patterns.

\rv{Of course, it is challenging to predict all possible uses of our work by others.
Some less obvious applications might include}
analysis of pattern vectors extracted from games of Go-playing programs:
the strength and style \rv{classification} might help to highlight some weaknesses
and room for improvements.%
\footnote{Of course, correlation does not imply causation \rv{and we certainly do not
suggest simply optimizing Go-playing programs according to these vectors.
However, they could hint on general shortcomings of the playing engines if the
actual cause of e.g. surprisingly low strength prediction is investigated.}}
Also, new historical game records are still \rv{occassionally} being discovered;
pattern-based classification might help to suggest \rv{or verify} origin of the games
in Go Archeology.

\section{Future Research}
\label{future-research}

Since we are not aware of any previous research on this topic and we
are limited by space and time constraints, plenty of research remains
to be done in all parts of our analysis --- we have already noted
many in the text above. Most significantly, different methods of generating
and normalizing the $\vec p$ vectors can be explored
and other data mining methods could be investigated.
Better ways of visualising the relationships would be desirable,
together with thorough expert dissemination of internal structure
of the player pattern vectors space:
more professional players should be consulted on the findings
and for style scales calibration.

It can be argued that many players adjust their style by game conditions
(Go development era, handicap, komi and color, time limits, opponent)
or that styles might express differently in various game stages\rvvv{.
These} factors should be explored by building pattern vectors more
carefully than by simply considering all moves in all games of a player.
Impact of handicap and uneven games on by-strength
$\vec p$ distribution should be also investigated.


\section{Conclusion}
We have proposed a way to extract summary pattern information from
game collections and combined this with various data mining methods
to show correspondence of our pattern summaries with various player
meta-information \rvvv{such as} playing strength, era of play or playing style,
as ranked by expert players. We have implemented and measured our
proposals in two case studies: per-rank characteristics of amateur
players and per-player style/era characteristics of well-known
professionals.

While many details remain to be worked out,
we have demonstrated that many significant correlations \rv{doubtlessly}
do exist and
it is practically viable to infer the player meta-information from
extracted pattern summaries \rv{and we have proposed applications}
for such inference. Finally, we outlined some of the many possible
directions of future work in this newly staked research field
on the boundary of Computer Go, Data Mining and Go Theory.


%

%

\section*{Acknowledgment}
\label{acknowledgement}

Foremostly, we are very grateful for detailed input on specific Go styles
by Alexander Dinerstein, Motoki Noguchi and V\'{i}t Brunner.
We appreciate helpful comments on our general methodology
by John Fairbairn, T. M. Hall, Cyril H\"oschl, Robert Jasiek, Franti\v{s}ek Mr\'{a}z
and several GoDiscussions.com users \cite{GoDiscThread}.
Finally, we would like to thank Radka ``chidori'' Hane\v{c}kov\'{a}
for the original research idea and acknowledge major inspiration
by R\'{e}mi Coulom's paper \cite{PatElo} on the extraction of pattern information.

\ifCLASSOPTIONcaptionsoff
  \newpage
\fi



\bibliographystyle{IEEEtran}
\bibliography{gostyle}
%
%
%

%






\end{document}